# Fuzzy Model on Human Emotions Recognition

KAVEH BAKHTIYARI &HAFIZAH HUSAIN
Department of Electrical, Electronics and Systems Engineering
Faculty of Engineering and Built Environment,
Universiti Kebangsaan Malaysia
43600 UKM Bangi, Selangor
MALAYSIA
hafizah.ukm@gmail.com

*Abstract:* - This paper discusses a fuzzy model for multi-level human emotions recognition by computer systems through keyboard keystrokes, mouse and touch-screen interactions. This model can also be used to detect the other possible emotions at the time of recognition. Accuracy measurements of human emotions by the fuzzy model are discussed through two methods; the first is accuracy analysis and the second is false positive rate analysis. This fuzzy model detects more emotions, but on the other hand, for some of emotions, a lower accuracy was obtained with the comparison with the non-fuzzy human emotions detection methods. This system was trained and tested by Support Vector Machine (SVM) to recognize the users' emotions. Overall, this model represents a closer similarity between human brain detection of emotions and computer systems.

## 1 Introduction

Computer systems are now attempting to interact more naturally with the users as human beings. Graphical User Interface (GUI) is becoming more flexible and intelligent to be adopted with human interests. The systems applications are willing to learn the user interactions to make a user friendly platform. Human emotions are the other user parameters which are being considered in a wide area.

The importance of human emotions recognition is reflected by its wide applications of this system in different areas. Computer games (entertainments), computer-based tutors (e-learning), machine and system controlling (industrial areas), criminology, computer privacy and security and so on [1].

For instance, the emotional states of students at a presentation in an online tutor reflect the usability and performance of the class on students. This information even can help teachers or intelligent systems to adapt the training system for a better educational ambiance to deliver more knowledge as possible to students [1, 2].

Recently, the accuracy of human emotions recognition is improving by employing different available methods and techniques such as image processing, voice recognition, natural language processing, keystroke dynamics, mouse movements, touch-screen interactions, EEG devices, measuring heart rates, body heat, blood pressure and so on [3-7]. Some methods like image processing, voice recognition and NLP are time and resource consuming and they need some special circumstances such as availability of video camera (webcam) or microphone, and some other methods require highly technical equipment such as for EEG or heart rate measuring [2]. These are medical machines which need a technical training to be installed and used, and sometimes the part of skin for installation should be shaved because of the inferences and noises. Keyboard keystroke dynamics, mouse movement and touch-screen interaction are the three methods which are mostly available on every computer or even portable digital devices such as handhelds, tablets, and mobile phones. This research has focused on these three methods for human emotions recognition because of availability and popularity of the devices and equipment.

Emotions are discussed by two factors in psychology. The first factor is Arousal which talks about the amount of energy of emotions. In literature, different amount of energy has been named such as happiness, sadness and so on. The second factor is Valence which shows the pleasure level, whether it is a positive or a negative emotion. There are 3 problems which are usually ignored in most of the researches for human emotions recognition in computer systems. Firstly, human emotions are not constant and they do not occur and exist in one single level. For instance, if you feel happiness for 2 days, the level and strength of feeling would not be the same for every moment that you have the same emotion. It means that emotions are more than being represented in 0 and 1. Nowadays, most of research papers present detection methods of emotion whether they exist or not. For example, if the user is happy or not; but it does not define the level of possible happiness. All the eight basic emotions according to the Sanskrit texts and the other sub-emotions are following the level based definitions [8]. Emotions have a fuzzy basis, and a discrete calculation would not be a solution on it.

The second problem is that users as human beings may not have only one emotion at a time; they may have different emotions in different levels. This second fact is also ignored in the most of researches, and the current human emotions recognition systems attempt to find only one current emotion and perhaps the strongest emotion at that time [9]. It is very clear that many emotions influence our life style and we are living with the combination of emotions. The happiness at the time of receiving a gift and the happiness at the time being in a theme park can be at a same level, but they are presented in 2 different qualities because the other available emotions are different in those two cases [10].

Thirdly, psychological studies show a difference in human emotional changes pattern based on cultural and language backgrounds. These differences say that a specific emotional pattern which is extracted from a specific area cannot be extended to the other part of the world [11].

This paper tries to analyze human emotions based on a fuzzy model which is able to recognize the human emotions in 5 levels and also to estimate the other possible emotions and their levels.

# 2 Methodology

Emotions' fuzzy model classifies the emotions into 5 different levels from 0 to 4. Zero (0) shows 0% of the emotion or in the other hand, it does not exist, and 4 shows the highest level of that appropriate emotion. Experience Sampling Methodology (ESM) has been used to collect the raw data for modeling classification. PANAS is a standard to find negative or positive emotions of the people [12]. PANAS uses 20 emotions and the seven missed basic emotions were added to this list in our research; and this standard was used to evaluate our system. The list of these 27 emotions is as follows:

| | | | |
|---|---|---|---|
| Joy | Surprise | Excited | Enthusias |
| Inspired | Active | Anticipatio | Fear |
| Upset | Proud | Nervous | Afraid |
| Anger | Acceptanc | Strong | Irritable |
| Determine | Disgust | Interested | Guilty |
| Alert | Attentive | Sadness | Distresse |
| Hostile | Ashamed | Jittery | |

This system was prepared as part of human emotions recognition by using keyboard keystroke dynamics, mouse movement, and touch screen interactions. Prototype software was developed based on Microsoft Windows to collect the required data, and users were prompted to enter their emotions and their level from 0 to 4 every 4 hours.

About 130 people participated in our project to have their emotional states recorded in our system. The people who participated were as below:

Men: 51% - Average Age: 30.13 Years Old
Women: 49% - Average Age: 28.29 Years Old

People were originally from:
Europe: 18.37%
Middle East: 44.18%
South East Asia: 13.17%
East Asia: 7.75%
People were living in:
Europe: 27.90%
South East Asia: 34.88%
East Asia: 16.37%
Middle East: 11.62%

45% of people were living in different regions of their own country.

In the collected data, men's emotions (self-reported) were stronger than women's, or at least they felt them stronger. The strongest emotion was *ACTIVE* in a range of 3.24 out of 4 and the least one was *UNFRIENDLY* with the value of 1.62 out of 4. It showed that most of the people thought that they were super active and less unfriendly in their normal life.

## 2.1 Analysis of Emotions

Here, only eight basic emotions were discussed which are *Joy*, *Anticipation*, *Anger*, *Disgust*, *Sadness*, *Surprise*, *Fear*, and *Acceptance*.

However, some similar change patterns were found among some emotions. Basically, both negative and positive emotions have similar change patterns, but between those basic emotions, some of them are very close to each other. For instance *Disgust* and *Anger* have a similar change rate pattern with each other, and *Sadness* is closer to them in comparison with the other emotions. On the other hand, *Acceptance* and *Anticipation* have very similar change pattern with each other. As it can be seen, these emotions are in a same group in terms of positive and negative emotions. Table 1 shows a similar change pattern, when the level of *Joy* is increasing from 0 to 4. Fig.1 also presents the graphed data of table 1.

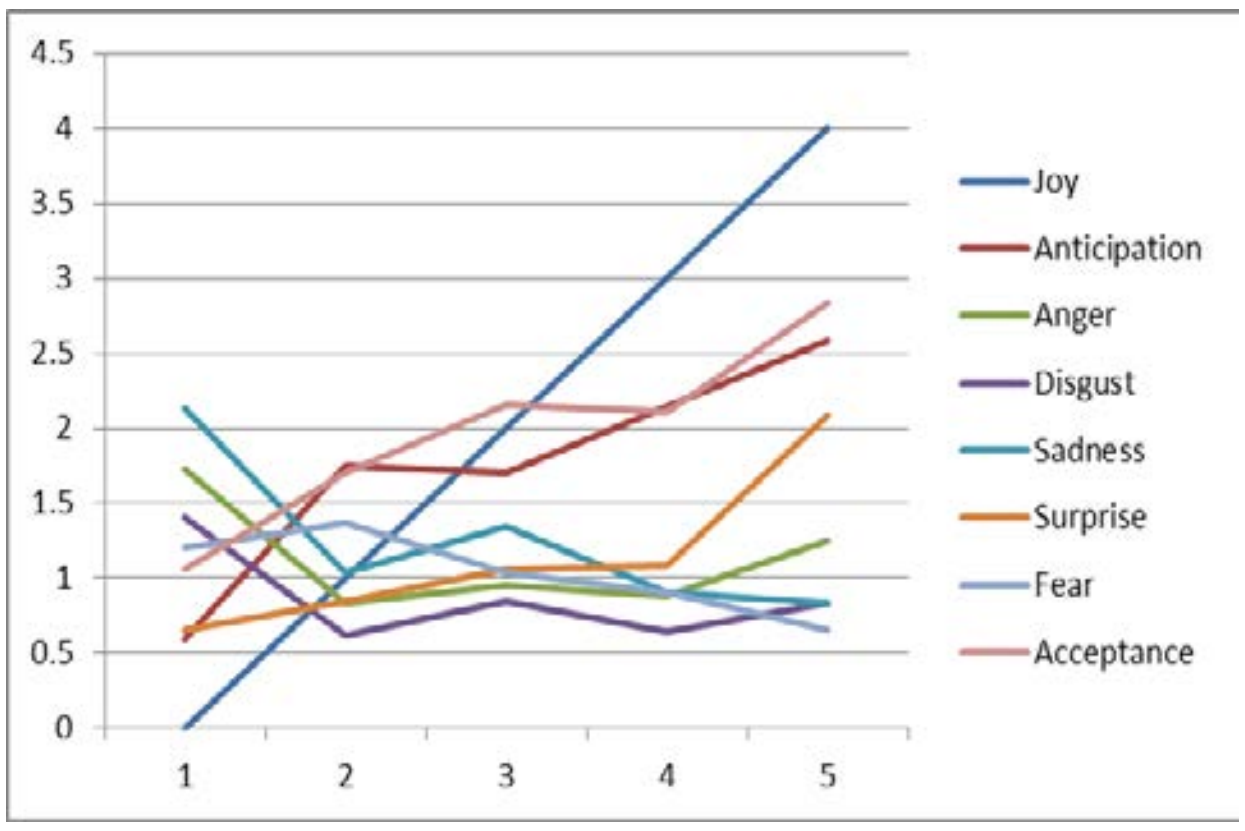

Fig.1. Emotions change pattern by the term of “Joy”

Conversely, the above values demonstrate the probabilistic of the existence for the other emotions. For example, when *Joy* is about 50%, it can be estimated that *Disgust* may exist at the level of 21%, or even *Fear* at the level of 26% can be possible; but we can be sure that 55% of *Fear* is unacceptable but *Acceptance* can be accepted in this level at the same time of 50% of *Joy*

The following 4 tables show the extracted analyzed data from our research based on the growth of some important basic emotions.

All above values are between 0 and 4. Later, it is tried to compare emotions changes pattern of people in different regions of the world. As described in the introduction, there are some psychological studies that indicate different emotions' concepts in different parts of the world [11, 13]. We have used our research data to confirm if the different emotions' concepts may cause different emotions changes patterns. For this section, 3 regions namely Europe, Middle East and South East Asia were chosen and the same data of the users with the emotion of *Joy* at the value of 2 or 50% were analyzed. Table 6 shows the classified values by regions.

Fig.2 demonstrates the graphed values in table 6. The emotional pattern of Europe and Middle East are similar, but South East Asia follows a very different pattern in comparison with the other two. This graph proves the psychological studies based on the influences of cultural and language backgrounds on emotions.

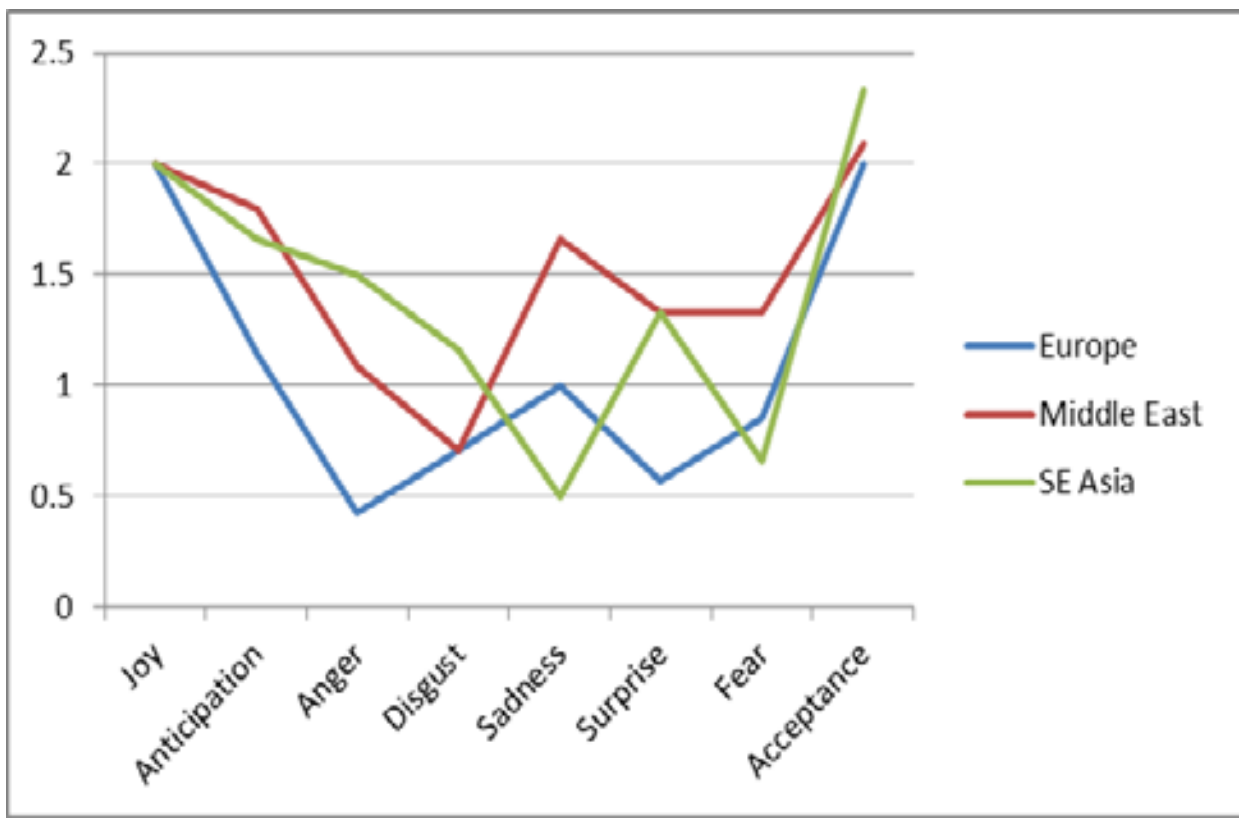

Fig.2. Classified graph of *Joy – 50%* for different regions

Table 1. Emotions change pattern by the term of *Joy*

| Anticipation | Anger | Disgust | Sadness | Surprise | Fear | Acceptance |
|---|---|---|---|---|---|---|
| 0.6 | 1.73 | 1.4 | 2.13 | 0.66 | 1.2 | 1.06 |
| 1.75 | 0.83 | 0.62 | 1.04 | 0.85 | 1.37 | 1.7 |
| 1.7 | 0.95 | 0.84 | 1.34 | 1.06 | 1.04 | 2.15 |
| 2.14 | 0.88 | 0.64 | 0.91 | 1.08 | 0.91 | 2.11 |
| 2.58 | 1.25 | 0.83 | 0.83 | 2.08 | 0.66 | 2.83 |

Table 2. Emotions change pattern based on the growth of *Anticipation*

| **Joy** | **Anger** | **Disgust** | **Sadness** | **Surprise** | **Fear** | **Acceptance** |
|---|---|---|---|---|---|---|
| **1.42** | 0.6 | 0.6 | 1.21 | 0.25 | 0.92 | 1.5 |
| **1.92** | 0.92 | 0.92 | 0.92 | 1.36 | 1.04 | 1.96 |
| **1.97** | 1.2 | 0.88 | 1.26 | 1.05 | 1.23 | 1.97 |
| **2.41** | 1.22 | 0.74 | 1.19 | 1.54 | 0.9 | 2.38 |
| **2.9** | 1.27 | 1.09 | 1.81 | 1.36 | 1.27 | 2.36 |

Table 3. Emotions change pattern based on the growth of “Anger”

| Joy | Anticipation | Disgust | Sadness | Surprise | Fear | Acceptance |
|---|---|---|---|---|---|---|
| **2.01** | 1.42 | 0.21 | 0.65 | 0.59 | 0.63 | 1.8 |
| **2.31** | 2 | 0.86 | 1 | 1.62 | 1.2 | 2.13 |
| **1.85** | 2.42 | 1.23 | 2.09 | 1.42 | 1.66 | 2.42 |
| **2.3** | 2.1 | 1.7 | 1.8 | 1.4 | 2 | 2 |
| **1.25** | 1.62 | 3 | 3.25 | 1.62 | 0.85 | 1.87 |

Table 4. Emotions change pattern based on the growth of *Fear*

| Joy | Anticipation | Anger | Disgust | Sadness | Surprise | Acceptance |
|---|---|---|---|---|---|---|
| **2.09** | 1.72 | 0.77 | 0.61 | 0.87 | 0.77 | 1.85 |
| **2.15** | 1.71 | 0.79 | 0.74 | 1.23 | 1.15 | 2.12 |
| **2.1** | 2.05 | 1.78 | 1.05 | 1.94 | 1.52 | 2.1 |
| **1.55** | 2.11 | 1.11 | 1.22 | 1.33 | 1.22 | 2 |
| **1.37** | 1.5 | 2 | 1.5 | 1.62 | 1.62 | 2.12 |

Table 5. Emotions change pattern based on the growth of *Acceptance*

| Joy | Anticipation | Anger | Disgust | Sadness | Surprise | Fear |
|---|---|---|---|---|---|---|
| **1.48** | 1.04 | 0.52 | 0.32 | 0.84 | 0.32 | 0.72 |
| **1.72** | 1.83 | 1.61 | 0.94 | 1.66 | 1.22 | 1.27 |
| **1.96** | 2.06 | 1 | 1.03 | 1.06 | 1.29 | 1.12 |
| **2.42** | 1.9 | 1.11 | 0.92 | 1.4 | 1.14 | 1.14 |
| **2.38** | 2.07 | 1 | 0.69 | 1.07 | 1.69 | 0.92 |

Table 6. Classified values of *Joy – 50%* for different regions

| | Europe | Middle | SE Asia |
|---|---|---|---|
| **Anticipation** | 1.14 | 1.8 | 1.66 |
| **Anger** | 0.42 | 1.09 | 1.5 |
| **Disgust** | 0.71 | 0.71 | 1.16 |
| **Sadness** | 1 | 1.66 | 0.5 |
| **Surprise** | 0.57 | 1.33 | 1.33 |
| **Fear** | 0.85 | 1.33 | 0.66 |
| **Acceptance** | 2 | 2.09 | 2.33 |

### 2.2 Fuzzy Human Emotions Recognition

In order to recognize the emotions, the system was trained and tested by Support Vector Machine (SVM) to recognize the users' emotions. The above results are now used in the evaluation for multi-level detection of emotions from 0 to 4, and also it is tried to recognize multi emotions at each stage of recognition process. Table 7 shows the final result of recognition after applying the fuzzy logic concept to detect each emotion by SVM in a range of 0 to 4.

The following table shows a lower accuracy in recognition of some of multi-level emotions in compare to the other non-fuzzy methods, because the number of available states has been increased to detect the emotion; but the accuracy in the emotion of *Nervous* has been increased.

Table 7. Confusion matrix with the average values

| **Selected Emotion** | **Detected Emotion** | | | |
|---|---|---|---|---|
| | Neutral | Afraid | Sadness | Nervous |
| **Neutral** | 0.68 | - | - | - |
| **Afraid** | - | 0.87 | - | - |
| **Sadness** | - | - | 0.86 | - |
| **Nervous** | - | - | - | 0.65 |

Application of fuzzy model on human emotions recognition systems represent more natural report on human emotions which includes all possible emotions and their levels and strengths. This report may not be as accurate as single emotion recognition like we had in the classic researches but it provides more reliable result for computer applications.

### 2.3 Fuzzy Model on Human Emotions Recognition: Discussion

The above data and classification were used as a fuzzy model in our human emotions recognition system. The accuracy of the emotions detection was evaluated from 2 different aspects. The first aspect evaluated whether the name of the recognized emotion or emotions are correct or not, and in the second aspect, it evaluated the system by the percentage of detected emotions [14, 15]. For example, consider a user with 2 emotions of Joy – 50% and *Fear* – 20%. In the first aspect of evaluation, if the system could detect *Joy* and *Fear*, evaluation is successfully done; but in the second aspect, not only *Joy* and *Fear* should be detected correctly, but also the detected level of the other emotions would be considered in the accuracy of the system [16].

This model was tested on human emotions recognition system which was working based on keyboard keystroke dynamics, mouse movements and touch-screen interactions. Support Vector Machine (SVM) was chosen as the system machine learning method.

After testing this fuzzy model on the system, the accuracy of the system from the first aspect increased from 3% to 5% as it was dependent on the emotions. This model estimated other emotions which can be possible in parallel with the other emotions.

From the second aspect of our evaluation, we gained the result from 0% to 16.7% of false positive rate for detecting the emotions level.

Both methods and aspects of evaluations for our analyzed data were reliable due to the high accuracy in the first method and low false positive rate for the second method.

## 3 Conclusion

This research proposed a new fuzzy based model through presenting a new solution based on the statistics and probabilistic for human emotions recognition to overcome some of the problems in this research area. This model detected the other possible emotions simultaneously, and worked more naturally by recognition of emotions in terms of levels and not through a discrete name of the emotion. The human emotions recognition system, which this model was implemented on, could gain up to 5% higher accuracy in the detection procedure, and a maximum 16.7% of false positive rate for recognizing the level of the emotions.

This model can be extended, improved, and even integrated into different systems. Future research work will be on human emotions recognition by extracted model for each region of the world and then comparing the accuracies and false positive rates of these different regions.

## Acknowledgement

We thank the students of Universiti Kebangsaan Malaysia (The National University of Malaysia) and Duisburg-Essen University. This research was part of a master's thesis of Artificial Intelligence in The National University of Malaysia.